%% file: 2025_ICCV_JYJUN.tex
\documentclass[10pt,twocolumn,letterpaper]{article}

\usepackage[pagenumbers]{iccv} 

\input{preamble}

\usepackage{multirow}

\definecolor{iccvblue}{rgb}{0.21,0.49,0.74}
\usepackage[pagebackref,breaklinks,colorlinks]{hyperref}

\newcommand{\bD}{\ensuremath{{\mathbf{D}}}}
\newcommand{\bG}{\ensuremath{{\mathbf{G}}}}
\newcommand{\bR}{\ensuremath{{\mathbf{R}}}}

\newcommand{\bI}{\ensuremath{{\mathbf{I}}}}

\newcommand{\bM}{\ensuremath{{\mathbf{M}}}}
\newcommand{\bN}{\ensuremath{{\mathbf{N}}}}

\usepackage[capitalize]{cleveref}
\crefname{section}{Sec.}{Secs.}
\Crefname{section}{Section}{Sections}
\Crefname{table}{Table}{Tables}
\crefname{table}{Tab.}{Tabs.}

\def\paperID{1074} 
\def\confName{ICCV}
\def\confYear{2025}

\title{Perfecting Depth: Uncertainty-Aware Enhancement of Metric Depth}

\author{
Jinyoung Jun\textsuperscript{1,2}\thanks{Work done during internship at Microsoft Research Asia.}\quad
Lei Chu\textsuperscript{2} \quad
Jiahao Li\textsuperscript{2} \quad
Yan Lu\textsuperscript{2} \quad
Chang-Su Kim\textsuperscript{1} \\[0.3em]
\textsuperscript{1}Korea University \quad
\textsuperscript{2}Microsoft Research Asia\\
{\tt\small jyjun@mcl.korea.ac.kr} \quad
{\tt\small \{leichu, jiahali, yanlu\}@microsoft.com} \quad
{\tt\small changsukim@korea.ac.kr}
}

\begin{document}
\maketitle

\begin{abstract}
We propose a novel two-stage framework for sensor depth enhancement, called Perfecting Depth. This framework leverages the stochastic nature of diffusion models to automatically detect unreliable depth regions while preserving geometric cues. In the first stage (stochastic estimation), the method identifies unreliable measurements and infers geometric structure by leveraging a training–inference domain gap. In the second stage (deterministic refinement), it enforces structural consistency and pixel-level accuracy using the uncertainty map derived from the first stage. By combining stochastic uncertainty modeling with deterministic refinement, our method yields dense, artifact-free depth maps with improved reliability. Experimental results demonstrate its effectiveness across diverse real-world scenarios. Furthermore, theoretical analysis, various experiments, and qualitative visualizations validate its robustness and scalability. Our framework sets a new baseline for sensor depth enhancement, with potential applications in autonomous driving, robotics, and immersive technologies.
\end{abstract}

\input{src/intro}
\input{src/related_work}
\input{src/method}
\input{src/experiments}
\input{src/conclusion}

{
    \small
    \bibliographystyle{ieeenat_fullname}
    \bibliography{2025_ICCV_JYJUN}
}

\onecolumn
\appendix

\input{src_supp/theory}

\input{src_supp/training_details}
\input{src_supp/sensor_depth_enhancement}
\input{src_supp/noisy_depth_completion}
\input{src_supp/depth_inpainting}
\input{src_supp/ethics}
\clearpage


\end{document}

%% file: preamble.tex
\usepackage[dvipsnames]{xcolor}


%% file: src/intro.tex
\section{Introduction}
\label{sec:introduction}
Depth maps represent pixel-wise distances from scene points to the camera, providing a fundamental representation of 3D structure. They play a crucial role in various 3D tasks, such as 2D-to-3D conversion \cite{Xie2016ECCV}, 3D model generation \cite{Izadinia2017CVPR}, and augmented reality \cite{du2020depthlab}. With the increasing availability of depth-capturing devices, dense and accurate depth maps have become essential in fields like autonomous driving, robotics, and immersive technologies.

Following breakthroughs in deep learning, monocular depth estimation has emerged as a practical alternative when depth sensors are unavailable. Early approaches focused on metric depth estimation \cite{eigen2014depth, lee2018single, lee2019big, fu2018deep, bhat2021adabins}. However, they rely on specific camera datasets \cite{silberman2012indoor, geiger2012we}, which limits real-world applicability due to scale ambiguity. To address this issue, recent research has shifted to relative depth estimation, which predicts pixel-wise depth order rather than absolute values \cite{ranftl2020, ke2024repurposing, yang2024depth, fu2024geowizard}.

\begin{figure}[!t]
  \centering
   \includegraphics[width=\linewidth]{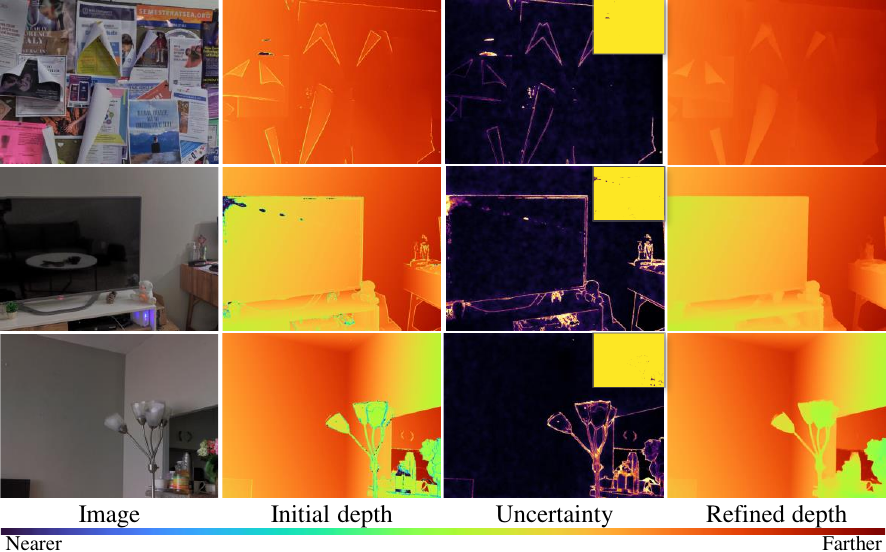}

   \vspace{-2mm}
   \caption{Sensor depth enhancement result on DIODE-Indoor. Without hand-crafted priors for noise or artifacts, our method effectively removes unreliable pixels and improves overall depth quality. Top-right inset indicates zero values in initial depth.}
   \vspace{-6mm}
   \label{fig:intro}
\end{figure}

While sufficient for some tasks, relative depth remains inadequate for applications requiring precise metric depth, such as high-fidelity 3D reconstruction or safety-critical robotic navigation. Moreover, sensor-based depth still faces challenges from hardware and environmental factors, leading to noise, missing data, and distortions—especially on reflective or transparent surfaces, as illustrated in Fig.~\ref{fig:intro}. These limitations hinder the reliability of depth maps in critical applications, necessitating more advanced approaches for depth enhancement.

To address the challenges of sensor depth enhancement, we propose a general framework, \textit{Perfecting Depth}. The key innovation of our approach lies in leveraging diffusion models to \textit{automatically} assess the reliability of each pixel through a carefully designed training-inference gap, thereby eliminating the reliance on manually defined artifact priors. Specifically, our framework consists of a two-stage pipeline: 
(1) \textit{Stochastic estimation} stage: We employ a diffusion-based strategy that trains the diffusion model on clean data but performs inference on raw, real-world data. This deliberate training-inference gap enables the model to measure per-pixel uncertainty and geometric cues, effectively identifying unreliable pixels without requiring handcrafted priors.
(2) \textit{Deterministic refinement} stage: A refinement network is then applied to focus on uncertain regions, ensuring accurate corrections while preserving valid sensor measurements.
By integrating the complementary strengths of stochastic sampling for global uncertainty estimation and precise local refinement for targeted corrections, our pipeline generates dense and high-quality depth maps that are well-suited for real-world tasks demanding accurate metric depth. We validate the effectiveness of our framework through theoretical analysis, extensive experiments, and qualitative visualizations, all of which demonstrate its superior performance.

Notably, the proposed framework is trained solely on synthetic data, yet it generalizes well to diverse real-world scenarios by inferring a global depth range from sensor input rather than relying on scale assumptions made during training. This underscores its potential for scalability across various real-world sensor datasets.

This paper has the following contributions:
\begin{itemize}
\itemsep0em
    \item We introduce a novel diffusion-deterministic pipeline for sensor depth enhancement, leveraging multiple reconstructions to detect noise without handcrafted priors.
    \item The proposed framework provides robust generalization to unseen, real-world sensor data even with synthetic-only training.
    \item The proposed framework excels in various scenarios, including sensor noise removal and depth completion with noisy measurements, surpassing state-of-the-art methods in real-world scenarios.
\end{itemize}

%% file: src/related_work.tex
\section{Related Work}
\label{sec:related_work}
\subsection{Sensor depth enhancement}
Depth maps from hardware sensors often suffer from imperfections due to sensor limitations. Various approaches have been explored to  effectively leverage sensor-derived depth, including depth completion, which reconstructs dense depth from sparse measurements \cite{ma2018sparse, cheng2018depth, cheng2019learning, park2020non, nazir2022semattnet, zhang2023completionformer, wang2023lrru, yu2023aggregating}, and depth super-resolution, which enhances low-resolution depth maps for better integration with high-resolution RGB data \cite{he2021towards, zhao2022discrete, zhao2023spherical, metzger2023guided}.  

However, these methods often assume that input depth maps are clean. While some depth completion methods address sparsity \cite{conti2023sparsity, jun2024masked, park2024depth}, handling real-world noise and inaccuracies remains a challenge. Similarly, depth super-resolution can amplify artifacts, especially in regions with complex geometries or sharp depth transitions.  

\subsection{Diffusion models}
Denoising Diffusion Probabilistic Model (DDPM) \cite{ho2020denoising} is a type of generative model which progressively removes Gaussian noise to approximate specific data distributions. Building on its superior training stability and generation quality, subsequent studies have focused on improving computational efficiency with Denoising Diffusion Implicit Model (DDIM) \cite{song2020denoising} and enabling high-resolution generation with Latent Diffusion Model (LDM) \cite{rombach2022high}. These advancements have significantly expanded the applicability of diffusion models across diverse fields, including image and video synthesis \cite{dhariwal2021diffusion, esser2023structure, blattmann2023align} and 3D generation \cite{bautista2022gaudi, zhang20233dshape2vecset}.

\subsection{Diffusion models with depth}
Diffusion models have been increasingly applied to depth-related tasks, primarily in monocular depth estimation \cite{ji2023ddp, saxena2024surprising, ke2024repurposing, tosi2024diffusion, patni2024ecodepth, fu2024geowizard, zhao2023unleashing}, where depth maps serve as noise while image representations provide conditioning \cite{ji2023ddp, saxena2024surprising, ke2024repurposing}. Another approach regenerates diverse images from a single depth map to address transparency and weather effects \cite{tosi2024diffusion}.  

Conversely, using depth maps to condition RGB generation has also been explored \cite{zhang2023adding, zhang2023jointnet, mo2024freecontrol, liu2024depthlab}, leveraging depth for structural guidance. However, these methods often struggle with precise alignment due to the stochastic nature of diffusion models. 

In contrast, our framework integrates stochastic sampling with deterministic refinement. We first generate multiple depth reconstructions using a diffusion-based masked autoencoder, then filter out unreliable pixels via multi-sample variance. The remaining reliable regions undergo deterministic refinement, ensuring structural consistency while enhancing depth with uncertainty awareness.  
\subsection{Masked autoencoders}
Masked Autoencoders (MAEs), inspired by masked language models \cite{devlin2018bert} and applied in vision transformers \cite{dosovitskiy2020image}, have proven effective for self-supervised learning and image reconstruction \cite{he2022masked, xie2022simmim}. MAEs are trained by masking parts of the input and predicting the missing regions, but they rely on fixed latent representations and feedforward architectures, making their outputs deterministic.
 
\begin{figure*}[!t]
  \centering
   \includegraphics[width=\linewidth]{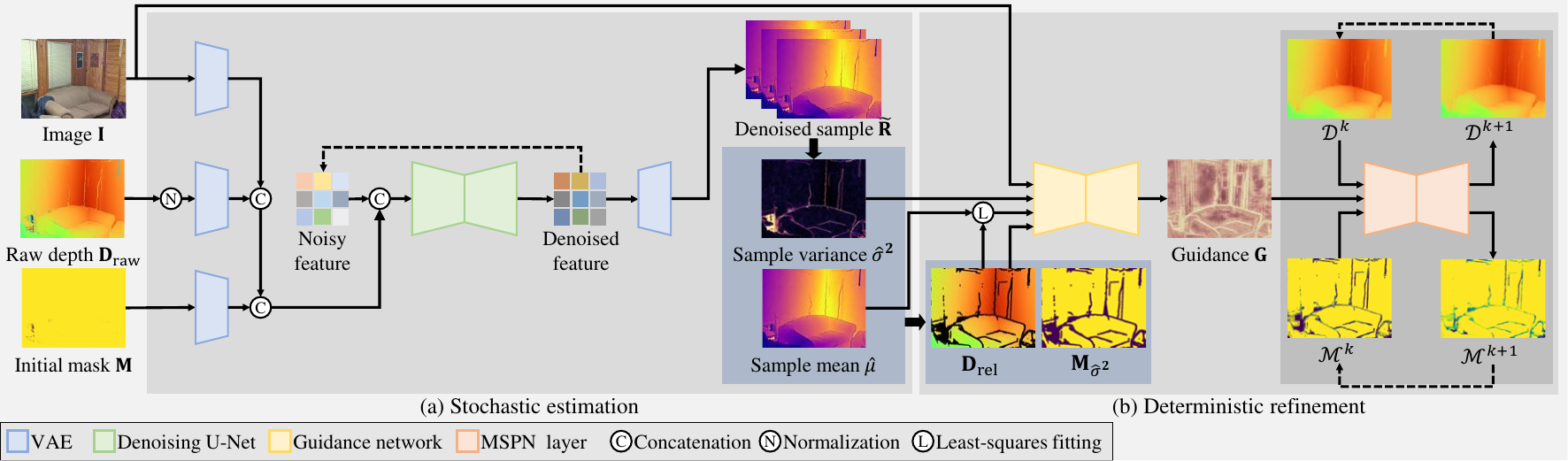}

   \vspace{-2mm}
   \caption{An overview of the proposed framework.}
   \vspace{-5mm}
   \label{fig:overview}
\end{figure*}

Our work extends MAEs by integrating RGB-D input for robust depth reconstruction. During training, the depth map is partially masked and reconstructed using RGB cues. We further incorporate diffusion models to inject random noise during reconstruction, allowing us to detect unreliable regions through increased variability across multiple reconstructions. This fusion of MAEs' structured encoding with the generative flexibility of diffusion models enables both unreliable region detection and inpainting for sensor depth enhancement under real-world conditions.
\subsection{Uncertainty in Neural Networks}
Uncertainty in neural networks is typically divided into epistemic uncertainty, reflecting model confidence in underrepresented regions, and aleatoric uncertainty, representing inherent data noise. Early works used Bayesian approximations \cite{gal2017deep, kendall2017uncertainties, kirsch2019batchbald} and ensembles \cite{lakshminarayanan2017simple, choi2018waic, chua2018deep, depeweg2018decomposition, postels2020hidden, berry2023escaping, berry2023normalizing, berry2024shedding} to capture epistemic uncertainty. Our approach, similar to concurrent work \cite{shu2024zero},  generates an ensemble through multiple denoising iterations of a single diffusion model, rather than by training multiple models. Different with these methods, the epistemic uncertainty of our pipeline is deliberately-introduced by the training-inference gap to approximate the  aleatoric uncertainty observed in real-world data, such as sensor artifacts or observational ambiguities.

%% file: src/method.tex
\section{Method}
\label{sec:method}

Given an RGB image $\bI \in \mathbb{R}^{3 \times H \times W}$ and its corresponding raw depth map $\bD_{\text{raw}} \in \mathbb{R}^{H \times W}$, our goal is to estimate a dense, artifact-free refined depth map $\bD_{\text{refine}}$ that accurately approximates the unknown ground-truth depth $\bD_{\text{true}}$. However, due to hardware limitations or external factors (e.g., reflections, sensor defects), certain pixels in $\bD_{\text{raw}}$ become unreliable. Formally, we write:
\begin{equation}
\bD_{\text{raw}} = \bD_{\text{true}} + \bN_{\text{sensor}},
\end{equation}
where $\bN_{\text{sensor}}$ denotes a set of localized sensor artifacts that do not follow a predefined distribution (e.g., Gaussian). This assumption reflects real-world scenarios where most depth values are accurate, but specific local regions may suffer severe corruption.

As illustrated in Fig.~\ref{fig:overview}, our approach consists of two main stages.
In Sec.~\ref{subsec:stochastic_estimation}, we introduce our novel approach that leverages the stochastic properties of the diffusion process to estimate the pixel-wise uncertainty and geometric cues via a carefully designed training-inference domain gap. In Sec.~\ref{subsec:deterministic_refinement}, we employ a deterministic network to refine high-uncertainty regions while preserving reliable measurements. This ensures that the final depth map is globally consistent and robust to input artifacts.

\subsection{Stochastic estimation}
\label{subsec:stochastic_estimation}
To avoid the need for manually defining the artifact prior and to enable automatic pixel-wise uncertainty estimation, we leverage the stochastic properties of the diffusion process. Specifically, we train a diffusion probabilistic model $\Theta$ to model the conditional distribution $p_{\Theta}$ as follows:
\begin{equation}
p_{\Theta}\bigl(\bD_\text{true}\,\big|\bI,\bD_\text{cond},\bM\bigr),
\label{eq:conditional_distribution_objective}
\end{equation}
where $\bM$ is a random mask and $\bD_{\text{cond}}\triangleq\bM\otimes\bD_{\text{true}}$ during training ($\otimes$ represents the Hadamard product). Rather than enumerating various types of artifacts during training, we train the diffusion network exclusively on clean data. Moreover, to prevent the model from overfitting to the metric range of the training data, we also apply scale-shift normalization that maps the depth values to the range $[-1, 1]$.

During inference, we condition the diffusion process on the raw depth map by setting $\bD_{\text{cond}} = \bD_{\text{raw}}$, where $\bM$ denotes the valid mask of non-zero values in $\bD_{\text{raw}}$. We will later show that this training–inference gap is key to estimating pixel reliability. We perform the denoising process $N$ times, starting from different random noise realizations, and obtain a set of denoised samples $\{\tilde{\bR}_i\}_{i=1}^N$. The pixel-wise expectation and variance of these samples are then utilized as both the geometric cues and uncertainty indicators for $\bD_{\text{raw}}$, respectively. Formally:
\begin{equation}
\Tilde{\bR}
\;\sim\;
p_{\Theta}\bigl(\bD_\text{true}\mid \bI,\bD_\text{cond},\bM\bigr).
\end{equation}
The pixel-wise posterior mean and variance are given by:
\begin{equation}
\mu(x,y)
\;=\;
\mathbb{E}_{p_\Theta}\!\bigl[\Tilde{\bR}(x,y)\bigr], 
\quad
\sigma^2(x,y)
\;=\;
\mathrm{Var}_{p_\Theta}\!\bigl[\Tilde{\bR}(x,y)\bigr].
\end{equation}
And we approximate them using $N$ samples as follows:
\begin{equation}
\begin{split}
\hat{\mu}(x,y) 
\;&=\;
\frac{1}{N} \sum_{i=1}^N \Tilde{\bR}^i(x,y),\\
\hat{\sigma}^2(x,y)
\;&=\;
\frac{1}{N}\sum_{i=1}^N \Bigl(\Tilde{\bR}^i(x,y)-\hat{\mu}(x,y)\Bigr)^2.
\label{eq:variance_est}
\end{split}
\end{equation}
In our implementation, we utilize synthetic data as the source of ground-truth depth maps, adapting the latent diffusion model as the underlying diffusion process. Details can be found in Sec.~\ref{subsec:implementation_details}.

\paragraph{Variance vs. Reliability.} In the Bayesian perspective, running the denoising process corresponds to drawing samples from the learned posterior distribution:  
\begin{equation}  
    p(\bD_{\text{true}} \mid \bI, \bD_{\text{cond}}, \bM),  
\end{equation}  
where $\bD_{\text{cond}} = \bD_{\text{raw}}$ during inference. At the pixel level, the posterior can be expressed as:  
\begin{equation}  
    p(\bD_{\text{true},(i,j)} \mid \bI, \bD_{\text{cond}}, \bM).  
\end{equation}  
Using Bayes' theorem, this posterior can approximated as:
\begin{equation}  
    \begin{split}  
        p(\bD_{\text{true},(i,j)} \mid \bI, \bD_{\text{cond}}&, \bM) \propto \\  
        p(\bD_{\text{cond},(i,j)} \mid &\bD_{\text{true},(i,j)}) p(\bD_{\text{true},(i,j)} \mid \bI, \bM). 
    \end{split}  
    \label{eq:bayesian_propo}
\end{equation}  
Here, the likelihood term $p(\bD_{\text{cond},(i,j)} \mid \bD_{\text{true},(i,j)})$ quantifies how well the input depth aligns with the ground-truth depth, whereas the prior term $p(\bD_{\text{true},(i,j)} \mid \bI, \bM)$ encodes the model's learned knowledge of plausible depth values based on the RGB image and the missing mask.  For detailed derivations, we refer readers to the appendix. 

In summary, Eq.~\ref{eq:bayesian_propo} suggests that (i) when the captured depth $\bD_{\text{raw}}$ is reliable (i.e., it aligns well with the training data), the likelihood term  becomes sharply peaked, yielding a narrow posterior and low posterior variance.  
(ii) Conversely, if the captured depth $\bD_{\text{raw}}$ is unreliable, the likelihood becomes weak, and the prior term  exerts more influence on the posterior. However, the prior alone may not provide sufficient information to fully resolve the uncertainty, resulting in a broader posterior and higher variance. Thus by identifying high-variance regions, we can selectively refine the most uncertain areas using a deterministic model, ensuring a more reliable depth reconstruction.

\noindent\textbf{Remarks.}   
In our formulation, we treat missing regions separately from other artifacts, as detecting missing areas is straightforward and computationally inexpensive. However, based on Eq.~\ref{eq:bayesian_propo}, both artifacts and missing areas share a common property: they exhibit weak likelihood terms and rely more heavily on the learned prior in the model. This observation implies that missing areas and artifacts behave similarly in terms of $\mu$ and $\sigma^2$. As a result, by filtering out unreliable depth measurements based on $\sigma^2$ during subsequent refinement, we primarily focus on the missing areas in the subsequent deterministic step. This simplification reduces the complexity of the training process, making it easier and more efficient to train the refinement network.

Furthermore, recent studies on depth have shown that excluding artifacts from real-world data brings synthetic datasets closer to their distribution~\cite{ke2024repurposing, liu2024depthlab}. This supports the idea that the likelihood term in Eq.~\ref{eq:bayesian_propo} aligns well with reliable regions, enabling the refinement network to be trained solely on synthetic data while still performing effectively in real-world scenarios.

\subsection{Deterministic refinement}
\label{subsec:deterministic_refinement}
Recall that in Sec~\ref{subsec:stochastic_estimation}, the depth map used during training is scale-shift normalized, so the actual metric range of the depth map is missing. In addition, because the diffusion process is global and stochastic, it is less effective for pixel-level corrections. Accordingly, we employ a deterministic refinement network to restore the metric range and further refine depth at the pixel-level.

Our refinement network is inspired by the sparsity-adaptive depth refinement (SDR) framework~\cite{jun2024masked}, which refines monocular depth estimates using sparse depth measurements. SDR utilizes the masked spatial propagation, where refinement starts from reliable depth values and progressively updates nearby pixels along with an evolving propagation mask. This ensures that reliable depth information is effectively propagated while refining unreliable depth estimates. We adopt this framework to complement our diffusion model by restoring absolute depth scales and improving pixel-level accuracy while preserving structural consistency.

We first filter out unreliable pixels using $\hat{\sigma}^2$ derived in Sec~\ref{subsec:stochastic_estimation}. A certainty mask $\bM_{\hat{\sigma}^2} \in \mathbb{R}^{H \times W}$ is defined as:
\begin{equation}  
\bM_{\hat{\sigma}^2}(x, y) =   
\begin{cases}   
0, & \text{if } \hat{\sigma}^2(x, y) > \epsilon, \\
1, & \text{otherwise}.  
\end{cases}
\label{eq:mask_sigma}
\end{equation}
Using this mask, we define a depth map with reliable values $\bD_{\text{rel}}$ and a scaled relative depth map $\bD_{\hat{\mu}}$ as follows:
\begin{equation}
\bD_{\text{rel}} = \bD_{\text{cond}} \otimes \bM \otimes \bM_{\hat{\sigma}^2},   
\quad  
\bD_{\hat{\mu}} =  a \cdot \hat{\mu} + b
\label{eq:D_rel_D_mu}
\end{equation}
where $\otimes$ denotes the Hadamard (element-wise) product. To remove small artifacts from hard thresholding, we apply morphological opening to $\bM_{\hat{\sigma}^2}$ before computing $\mathbf{D}_{\text{rel}}$. The scale and shift parameters $a$ and $b$ are obtained by solving the least-squares problem:  
\begin{equation}
\{a, b\} = \arg\min_{a, b} \|\bD_{\text{rel}} - (a \cdot \hat{\mu} + b) \|^2.
\label{eq:least_squares}
\end{equation}
Next, we extract a guidance feature $\bG$ using a feature extraction module $\Phi_{\text{guide}}$, which encodes both global context and local uncertainty cues. The guidance feature is formulated as:
\begin{equation}
\bG = \Phi_{\text{guide}}(\bI, \bD_{\text{rel}}, \bD_{\hat{\mu}}, \hat{\sigma}^2).
\label{eq:mspn_guide}
\end{equation}
In Eq.~\ref{eq:mspn_guide}, including $\bD_{\text{rel}}$, $\bD_{\hat{\mu}}$, and $\hat{\sigma}^2$ provides complementary information that enhances the refinement process. $\bD_{\text{rel}}$ retains only depth values deemed trustworthy based on the certainty mask, serving as a reliable reference. However, filtering removes noisy depth values at the cost of increased sparsity, which may disrupt global consistency.

On the other hand, $\bD_{\hat{\mu}}$ retains the global scene structure from the diffusion network while restoring absolute depth scales via least-squares fitting, ensuring geometric coherence. Additionally, $\hat{\sigma}^2$ encodes pixel-wise uncertainty, allowing the network to adaptively update weights during masked spatial propagation. By combining these three components, the guidance feature $\bG$ provides a strong structural prior with metric scale information while preserving both global consistency and local adaptability, thus ensuring effective depth refinement.

With $\bG$, we iteratively refine the depth map using the masked spatial propagation module $\Phi_{\text{MSPN}}$, as proposed in~\cite{jun2024masked}. The refinement process starts with the initial depth map $\mathcal{D}^0$ and mask $\mathcal{M}^0$ defined as:
\begin{equation}
\mathcal{D}^0 = \bD_{\text{rel}}, \quad \mathcal{M}^0 = \bM \otimes \bM_{\hat{\sigma}^2}.
\end{equation}
At each iteration $k$, the depth map $\mathcal{D}^k$ and mask $\mathcal{M}^k$ are updated as follows:
\begin{equation}  
(\mathcal{D}^{k+1}, \mathcal{M}^{k+1}) = \Phi_{\text{MSPN}}(\mathcal{D}^k, \mathcal{M}^k, \bG), \quad k = 0, \dots, K-1,  
\end{equation}
where $K$ is the total number of iterations. The final refined depth map is obtained as $\bD_{\text{refined}} = \mathcal{D}^K$.

This refinement pipeline applies localized corrections to uncertain regions identified by $\hat{\sigma}^2$, yielding $\bD_{\text{refined}}$ with improved structural coherence and robustness to sensor noise. By leveraging the complementary strengths of stochastic diffusion for global consistency and deterministic refinement for local accuracy, our two-stage framework effectively balances global plausibility with pixel-level precision, thereby mitigating the limitations inherent to each individual method.

To train our deterministic refinement network, we also use the outputs of our diffusion model also on synthetic data. For each sample, we randomly generate a mask $\bM$ and compute $\bD_{\text{cond}}$ from $\bD_{\text{true}}$. Then, we perform inference through the diffusion model to obtain $\hat{\mu}$ and $\hat{\sigma}^2$. We treat $(\bD_{\text{cond}}, \hat{\mu}, \hat{\sigma}^2)$ as inputs, and optimize the refinement network by minimizing the error against the known ground-truth depth $\bD_{\text{true}}$ using L1 and L2 loss, following~\cite{jun2024masked}. 

This training strategy enables the refinement network to leverage the posterior estimates of the stochastic diffusion model while still preserving metric accuracy through supervised learning on synthetic ground truth.

\noindent\textbf{Remarks.}  
To prevent the network from overfitting to the specific metric range in the training dataset, we want the network to infer the metric range directly from the raw depth map $\bD_{\text{raw}}$. To achieve this, we apply a random scale and shift to the depth data during training of the deterministic network. This strategy enhances the generalizability of our method to in real-world data, where the metric range may vary significantly. Furthermore, it facilitates the integration of multiple datasets with differing metric ranges, enabling the scalability of our model to larger and more diverse training datasets.

%% file: src/experiments.tex
\section{Experiments}
\label{sec:experiments}

\subsection{Implementation details}
\label{subsec:implementation_details}
\noindent\textbf{Training dataset:} 
We train our framework on the Hypersim dataset \cite{roberts2021hypersim}, a synthetic indoor dataset with dense per-pixel ground-truth depth $\bD_{\text{true}}$. This ground truth is essential for our formulation in Eq.~\ref{eq:conditional_distribution_objective}, where we construct $\bD_{\text{true}}$ by applying structured masking to $\bD_{\text{true}}$ during training.

Although a synthetic outdoor dataset, Virtual Kitti 2~\cite{cabon2020virtual}, can be used for training, it is designed for autonomous driving scenarios with lower variability. We therefore utilize Hypersim for training and evaluate the framework on indoor datasets.

\noindent\textbf{Evaluation dataset:} 
To evaluate the performance of the proposed framework, we adopt three commonly used real-world indoor datasets, DIODE-Indoor~\cite{vasiljevic2019diode}, NYUv2~\cite{silberman2012indoor}, and ScanNet~\cite{dai2017scannet}. DIODE-Indoor provides dense but noisy data. Meanwhile, NYUv2 and ScanNet provide relatively clean depth maps, but include missing regions. For all experiments, we set $\epsilon=0.01$.

\noindent\textbf{Diffusion network:}  
We fine-tune Stable Diffusion v2 \cite{rombach2022high}, a latent diffusion model trained on LAION-5B \cite{schuhmann2022laion}. The VAE remains frozen and encodes inputs, while the denoising U-Net is modified to accept a 16-channel input. The final depth reconstruction is obtained by averaging the decoded VAE output across channels.

\noindent\textbf{Refinement network:}  
For refinement, we employ MaxViT \cite{tu2022maxvit} as the feature extractor $\Phi_{\text{guide}}$ and utilize MSPN \cite{jun2024masked} for masked spatial propagation. Refinement is performed iteratively over six steps, with two MSPN layers in series (window sizes 13 and 3).



\begin{table}[!t]
    \centering
    \small
    \renewcommand{\arraystretch}{1.3}
    \setlength{\tabcolsep}{12pt}
    \begin{tabular}{l l}
    \toprule
    RMSE & $\frac{1}{|\bD|}\big(\sum_{i}(\hat{d}_i-d_i)^2\big)^{0.5}$\\
    \midrule
    $\delta_k$ & \% of $d_i$ that satisfies
    $\max\!\left\{\frac{\hat{d}_i}{d_i},\frac{d_i}{\hat{d}_i}\right\}<k$\\
    \midrule
    Kendall's $\tau$ & $\frac{\alpha(\hat{D}, D) - \beta(\hat{D}, D)}{\binom{|D|}{2}}$\\
    \bottomrule
    \end{tabular}
    \vspace{-3mm}
    \caption
    {
        Evaluation metrics for estimated depth maps. Here, $|\bD|$ denotes the number of valid pixels in a depth map $\bD$, $d_i$ is the $i$th valid depth in $\bD$, and $\hat{d}_i$ is an estimate of $d_i$. $\alpha(\cdot)$ and $\beta(\cdot)$ count the number of concordant and discordant depth pairs, respectively.
    }
    \label{tb:eval_metric}
    \vspace{-5mm}
\end{table}

\noindent\textbf{Training details:} More details are provided in appendix.

\begin{table}[!t]
    \scriptsize
    \setlength{\tabcolsep}{3.5pt}
    
    \centering
    \begin{tabular}{l|l|ccc|ccc}
    \toprule
    \multirow{2}{*}{Method} & \multirow{2}{*}{Train data} & \multicolumn{3}{c|}{DIODE-Indoor} & \multicolumn{3}{c}{NYUv2}\\
     & & RMSE$_\downarrow$ & $\delta_{1.25\uparrow}$ & $\tau_\uparrow$ & RMSE$_\downarrow$ & $\delta_{1.25\uparrow}$ & $\tau_\uparrow$\\
    \midrule
    \multirow{3}{*}{DA V1~\cite{yang2024depth}} & Raw data & 0.837 & 0.719 & 0.819 & 0.282 & 0.959 & 0.838\\
    & Diff.~only & 0.818 & 0.725 & 0.803 & 0.246 & 0.960 & 0.864\\
    \cmidrule{2-8}
    & Proposed & \textbf{0.816} & \textbf{0.727} & \textbf{0.820} & \textbf{0.223} & \textbf{0.974} & \textbf{0.873}\\
    \midrule
    \multirow{3}{*}{DA V2~\cite{depth_anything_v2}} & Raw data & 0.828 & 0.724 & 0.845& 0.263 & 0.965 & 0.846\\
    & Diff.~only & 0.821 & 0.719 & 0.821 & 0.249 & 0.955 & 0.867\\
    \cmidrule{2-8}
    & Proposed & \textbf{0.817} & \textbf{0.725} & \textbf{0.851} & \textbf{0.217} & \textbf{0.972} & \textbf{0.877}\\
    \bottomrule
    \end{tabular}
    \vspace{-3mm}
    \caption{Fine-tuning performance of relative depth estimators using DIODE-Indoor. $\tau$ denotes Kendall's $\tau$.}
    \label{tb:sensor_depth_enhancement}
    \vspace{-3mm}
\end{table}

\begin{figure}[!t]
  \centering
   \includegraphics[width=\linewidth]{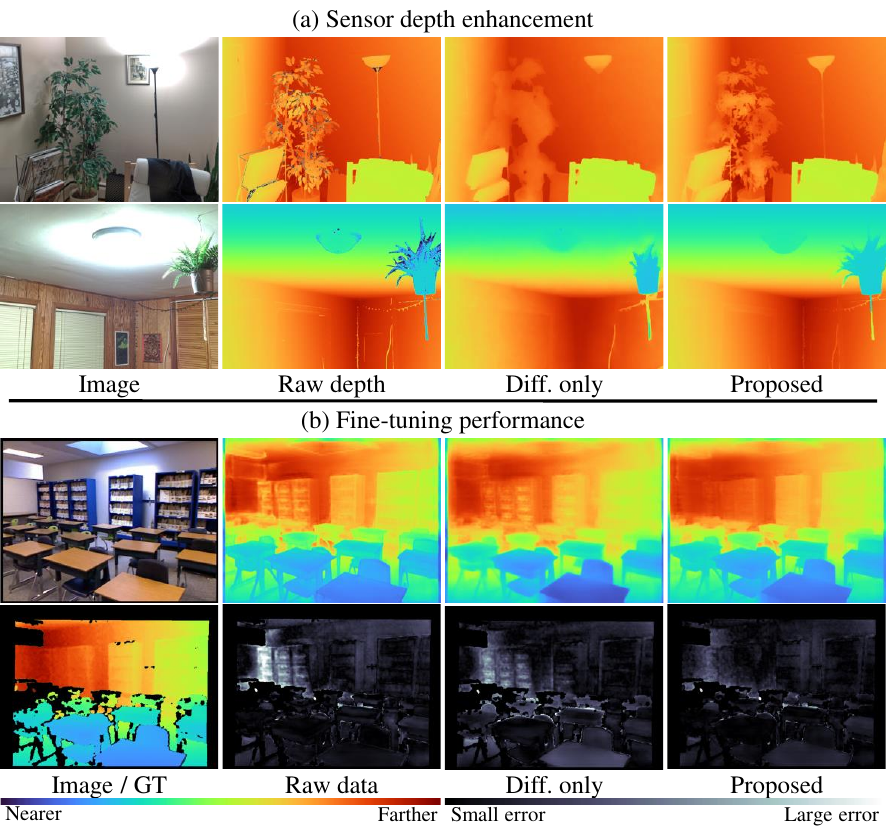}
   \vspace{-5mm}
   \caption{(a) Sensor depth enhancement result on DIODE-Indoor. (b) Qualitative comparison of fine-tuning results on NYUv2. For each depth map in (b), the corresponding error map is provided below, in which brighter pixels represent large errors.}
   \label{fig:qualitative_sensor_depth_enh}
    \vspace{-5mm}
\end{figure}

\subsection{Evaluation metrics}
As described in Tab.~\ref{tb:eval_metric}, we adopt two metrics to measure the quality of metric depth map. For the sensor depth enhancement experiment in Sec~\ref{subsec:sensor_depth_enhancement}, we additionally adopt Kendall's $\tau$ \cite{kendall1938new} to measure the accuracy for relative depth.

\subsection{Sensor depth enhancement}
\label{subsec:sensor_depth_enhancement}
We assess our sensor depth enhancement framework using the noisy DIODE-Indoor dataset \cite{vasiljevic2019diode}. Since obtaining ground-truth $\bD_{\text{true}}$ is often infeasible, direct evaluation on individual depth maps is challenging. Hence, we conduct an indirect evaluation by fine-tuning relative depth estimators on the refined depth maps and assessing their performance. For this experiment, we adopt Depth Anything (DA) V1 and V2 \cite{yang2024depth, depth_anything_v2}, neither of which was originally trained on DIODE-Indoor. We fine-tune each network five times and report the average performance.

To compare the effectiveness of our framework, we set three training data configurations. `Raw data' refers to the original sensor depth maps without any enhancement. `Diff.~only' first removes unreliable pixels using the uncertainty map from our stochastic estimation stage. The same diffusion model is then re-applied to inpaint those missing regions, followed by least squares fitting to recover the global scale as in Eq.~\ref{eq:least_squares}. `Proposed' represents the full pipeline, incorporating both stochastic estimation and deterministic refinement.

Since the test labels in DIODE-Indoor are noisy as well, we filter out unreliable pixels using $\hat{\sigma}^2$ from the first stage of our pipeline and measure metric depth performance. Additionally, we evaluate zero-shot relative depth estimation performance on the NYUv2 test set, a widely used benchmark for depth estimation. This dual evaluation confirms that our framework not only improves data quality in noisy regions but also generalizes well across datasets. The following observations can be made from Tab.~\ref{tb:sensor_depth_enhancement} and Fig.~\ref{fig:qualitative_sensor_depth_enh}:

\begin{table}[!t]
    \scriptsize
    \setlength{\tabcolsep}{4.9pt}
    
    \centering
    \begin{tabular}{l|cc|cc|cc}
    \toprule
    \multirow{2}{*}{Method} & \multicolumn{2}{c|}{5\%} & \multicolumn{2}{c|}{10\%} & \multicolumn{2}{c}{20\%} \\
    & RMSE$_\downarrow$ & $\delta_{1.25\uparrow}$ & RMSE$_\downarrow$ & $\delta_{1.25\uparrow}$ & RMSE$_\downarrow$ & $\delta_{1.25\uparrow}$\\
    \midrule
    CFormer~\cite{zhang2023completionformer} & 0.261 & \textbf{0.976} & 0.370 & 0.952 & 0.526 & 0.897\\
    + Proposed & \textbf{0.216} & 0.972 & \textbf{0.219} & \textbf{0.971} & \textbf{0.225} & \textbf{0.970}\\
    \midrule
    MSPN~\cite{jun2024masked} & 0.301 & \textbf{0.972} & 0.437 & 0.947 & 0.655 & 0.888\\
    + Proposed & \textbf{0.216} & \textbf{0.972} & \textbf{0.220} & \textbf{0.971} & \textbf{0.227} & \textbf{0.970}\\
    \bottomrule
    \end{tabular}
    \vspace{-3mm}
    \caption{Noisy depth completion and its refinement performance on NYUv2 using 500 sparse depths. \% denotes the noise ratio.}
    \label{tb:noisy_depth_completion}
    \vspace{-3mm}
\end{table}

\begin{figure}[!t]
  \centering
   \includegraphics[width=\linewidth]{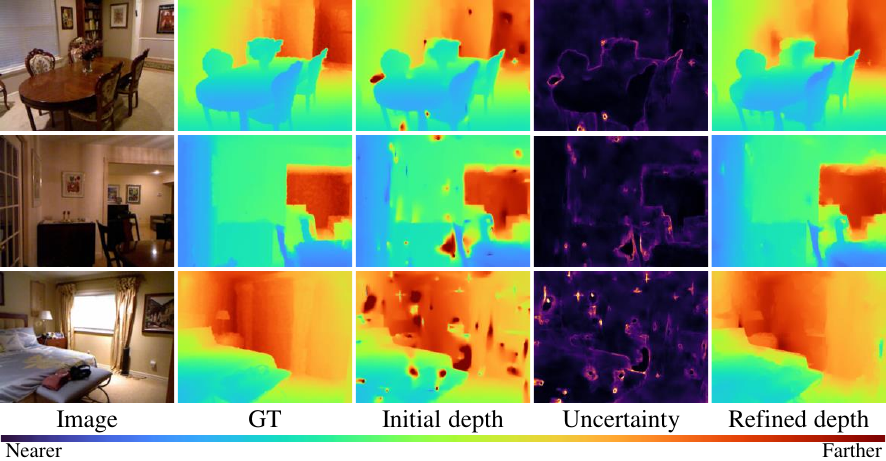}
   \vspace{-6mm}
   \caption{Refinement result on noisy depth completion using MSPN. From the first row, noise ratios are 5\%, 10\%, and 20\%, respectively.}
   \label{fig:qualitative_noisy_DC}
    \vspace{-6mm}
\end{figure}

\begin{table*}[!t]
    \scriptsize
    \setlength{\tabcolsep}{5.1pt}
    
    \centering
    \begin{tabular}{c|l|cc|cc|cc|cc|cc|cc}
    \toprule
    \multirow{2}{*}{Dataset} & \multirow{2}{*}{Method}
    & \multicolumn{2}{c|}{H2I $\in (0.01, 0.1]$}
    & \multicolumn{2}{c|}{H2I $\in (0.1, 0.2]$}
    & \multicolumn{2}{c|}{H2I $\in (0.2, 0.3]$}
    & \multicolumn{2}{c|}{H2I $\in (0.3, 0.4]$}
    & \multicolumn{2}{c|}{H2I $\in (0.4, 0.5]$}
    & \multicolumn{2}{c}{Average}\\
    & & RMSE$_\downarrow$ & $\delta_{1.25\uparrow}$
    & RMSE$_\downarrow$ & $\delta_{1.25\uparrow}$
    & RMSE$_\downarrow$ & $\delta_{1.25\uparrow}$
    & RMSE$_\downarrow$ & $\delta_{1.25\uparrow}$
    & RMSE$_\downarrow$ & $\delta_{1.25\uparrow}$
    & RMSE$_\downarrow$ & $\delta_{1.25\uparrow}$\\
    \midrule
    \multirow{4}{*}{NYUv2} & Depth Anything V1~\cite{yang2024depth}
    & 0.460 & 0.894
    & 0.396 & 0.869
    & 0.412 & 0.866
    & 0.417 & 0.871
    & 0.413 & 0.861
    & 0.420 & 0.872\\
    & Depth Anything V2~\cite{depth_anything_v2}
    & 0.460 & 0.887
    & 0.400 & 0.864
    & 0.418 & 0.861
    & 0.422 & 0.863
    & 0.417 & 0.853
    & 0.423 & 0.866\\
    & Marigold~\cite{ke2024repurposing}
    & 0.336 & 0.931
    & 0.296 & 0.919
    & 0.315 & 0.916
    & 0.315 & 0.921
    & \textbf{0.299} & 0.917
    & 0.312 & 0.921\\
    \cmidrule{2-14}
    & Proposed
    & \textbf{0.261} & \textbf{0.954}
    & \textbf{0.213} & \textbf{0.968}
    & \textbf{0.216} & \textbf{0.965}
    & \textbf{0.296} & \textbf{0.941}
    & 0.305 & \textbf{0.931}
    & \textbf{0.259} & \textbf{0.952}\\
    \midrule
    \multirow{4}{*}{ScanNet} & Depth Anything V1~\cite{yang2024depth}
    & 0.220 & 0.920
    & 0.210 & 0.916
    & 0.210 & 0.917
    & 0.208 & 0.917
    & 0.212 & 0.916
    & 0.212 & 0.917\\
    & Depth Anything V2~\cite{depth_anything_v2}
    & 0.224 & 0.919
    & 0.218 & 0.915
    & 0.217 & 0.915
    & 0.217 & 0.916
    & 0.219 & 0.913
    & 0.219 & 0.916\\
    & Marigold~\cite{ke2024repurposing}
    & 0.225 & 0.905
    & 0.216 & 0.902
    & 0.219 & 0.901
    & 0.220 & 0.910
    & 0.218 & 0.898
    & 0.220 & 0.903\\
    \cmidrule{2-14}
    & Proposed
    & \textbf{0.110} & \textbf{0.974}
    & \textbf{0.104} & \textbf{0.984}
    & \textbf{0.115} & \textbf{0.980}
    & \textbf{0.140} & \textbf{0.970}
    & \textbf{0.162} & \textbf{0.960}
    & \textbf{0.126} & \textbf{0.974}\\
    \bottomrule
    \end{tabular}
    \vspace{-2.5mm}
    \caption{Depth inpainting performance on NYUv2 and ScanNet according to the hole-to-image (H2I) area ratios.}
    \label{tb:depth_inpainting}
    \vspace{-2mm}
\end{table*}

\begin{figure*}[!t]
  \centering
   \includegraphics[width=\linewidth]{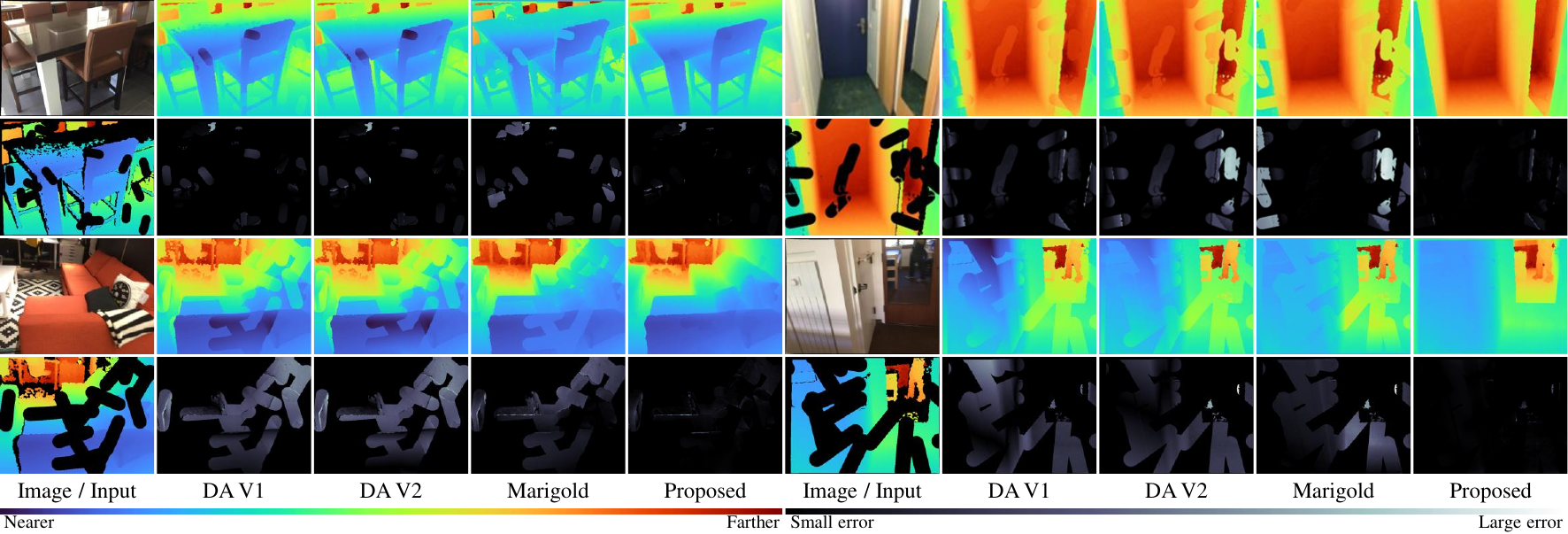}
   \vspace{-6mm}
   \caption{Qualitative comparison of depth inpainting results on ScanNet. For each depth map, the correponding error map is provided below, in which brighter pixels represent large errors.}
   \label{fig:qualitative_depth_inpainting}
    \vspace{-5mm}
\end{figure*}


\begin{itemize}
\itemsep0mm
\item In Tab.~\ref{tb:sensor_depth_enhancement}, we see that `Proposed' consistently outperforms its alternatives not only on DIODE-Indoor but also on NYUv2, demonstrating strong cross-dataset generalization. Specifically, for DA V2, `Proposed' improves RMSE and $\tau$ by 21.2\% and 4.2\%, respectively, compared with `Raw data.'
\itemsep0mm
\item As illustrated in Fig.~\ref{fig:qualitative_sensor_depth_enh} (a), `Raw data' contains noisy pixels, particularly around edge regions. `Diff.~only' suppresses such noise but often results in blurry refinements. In contrast, `Proposed' effectively reduces structured noise while preserving fine edge details.
\itemsep0mm
\item These characteristics are also reflected in the fine-tuned results, as illustrated in Fig.~\ref{fig:qualitative_sensor_depth_enh} (b). The improved depth quality leads to more accurate predictions in a downstream task.
\itemsep0mm
\item Notably, `Raw data' tends to learn incorrect depth values near edge regions, as seen in Fig.~\ref{fig:qualitative_sensor_depth_enh} (b). While fine-tuning reveals larger disparities in challenging edges, the performance gap in DIODE-Indoor appears smaller since noisy pixels are removed before evaluation. In contrast, for the cleaner NYUv2 dataset, `Proposed' achieves even greater improvements.
\itemsep0mm
\item Overall, our framework enhances depth quality by mitigating noise, thus effectively providing higher-quality depth maps for downstream tasks.
\itemsep0mm
\end{itemize}

\subsection{Noisy depth completion}
\label{subsec:noisy_depth_completion}
Although directly enhancing real sensor depth is challenging, we can still evaluate the robustness of our framework by injecting sensor-like noise in a controlled setting. To this end, we conduct a refinement experiment on noisy depth completion results, where our approach is applied to remove induced artifacts and improve depth quality.

Conventional depth completion methods generate dense maps from sparse, noise-free measurements. However, real-world sensor depth is often corrupted by noise due to hardware limitations and environmental factors, leading to artifacts that degrade depth accuracy. To assess how well different methods handle such artifacts, we simulate noisy depth completion conditions by introducing noise into the sparse depth inputs. Specifically, we randomly select valid pixels from the ground truth, designate a fraction (5\%, 10\%, and 20\%) as noisy, and add Gaussian noise while ensuring non-negative depth values.

Unlike synthetic depth datasets or artificially degraded clean data, our noisy depth simulation more closely reflects real-world sensor noise, as it introduces a diverse range of artifacts beyond simple pixel-wise corruption. This setting allows us to evaluate not only robustness to synthetic noise but also the ability to refine the structured sensor artifacts that may appear in real-world depth data.

For this experiment, we evaluate our approach alongside a depth completion method, CFormer \cite{zhang2023completionformer}. Additionally, we test MSPN \cite{jun2024masked}, an SDR framework, which leverages monocular depth estimation as a prior. Tab.~\ref{tb:noisy_depth_completion} and Fig.~\ref{fig:qualitative_noisy_DC} reveals the following observations:

\begin{itemize}
\itemsep0mm
\item In Tab.~\ref{tb:noisy_depth_completion}, our framework consistently improves the noisy depth completion across varying noise levels. Although trained solely on synthetic data, it generalizes well to real-world artifacts, making it a promising solution for sensor depth refinement.
\itemsep0mm
\item At higher noise ratios, conventional methods degrade significantly, whereas our framework maintains robustness, effectively suppressing noise in challenging conditions. For instance, it improves MSPN's RMSE performance by 289\%.
\itemsep0mm
\item In Fig.~\ref{fig:qualitative_noisy_DC}, our approach excels in structured regions, enhancing consistency and preserving geometric details. 
\itemsep0mm
\item These results also validate our designed train-inference domain gap in Sec.~\ref{subsec:stochastic_estimation}. This discrepancy enables uncertainty estimation and selective refinement, enhancing the adaptability to real-world sensor artifacts.
\itemsep0mm
\end{itemize}

\begin{figure}[!t]
  \centering
   \includegraphics[width=\linewidth]{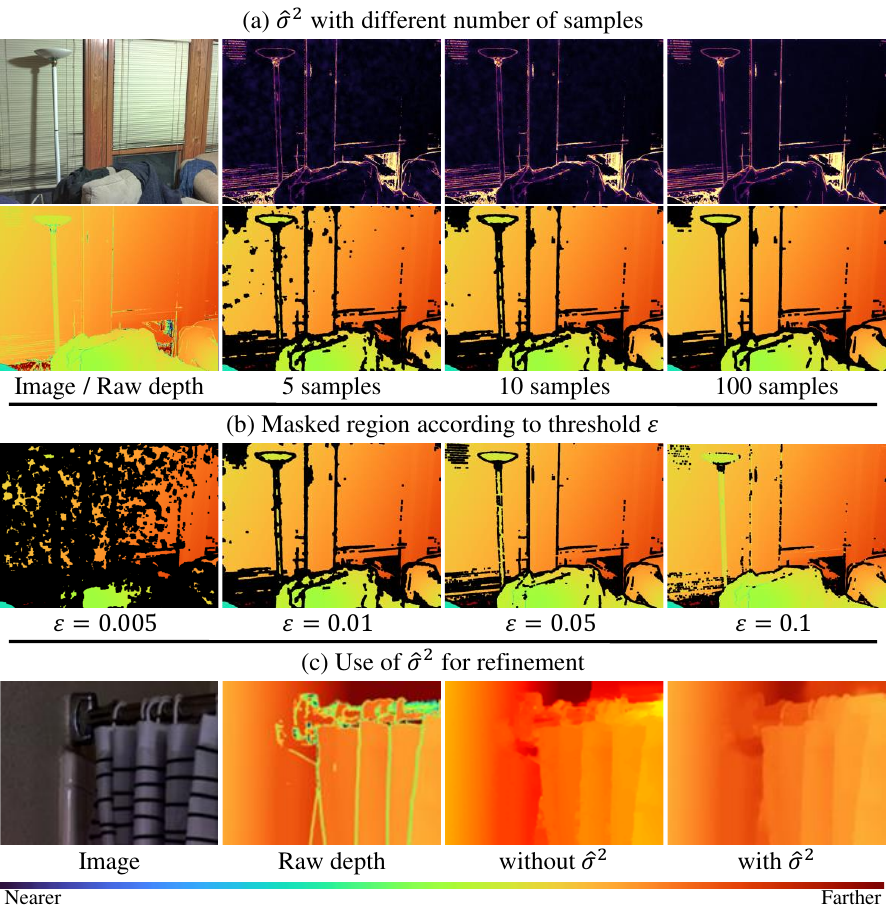}
   \vspace{-6mm}
   \caption{(a) Visualization of $\hat{\sigma}^2$ and the corresponding masked depth map according to $N$. (b) Masking result according to threshold $\epsilon$. (c) Efficacy of $\hat{\sigma}^2$ for refinement.}
   \vspace{-6mm}
   \label{fig:analysis}
\end{figure}

\subsection{Depth inpainting}
\label{subsec:depth_inpainting}
Our framework demonstrated strong performance in noisy depth maps in Sec.~\ref{subsec:sensor_depth_enhancement} and ~\ref{subsec:noisy_depth_completion}. However, handling depth maps with large missing area poses a different challenge. In such cases, one potential alternative is to leverage a monocular relative depth estimator, where known depth values can be used to recover the global scale of missing regions.

To assess this, we evaluate our diffusion network for inpainting against state-of-the-art monocular relative depth estimators \cite{yang2024depth, depth_anything_v2, ke2024repurposing}. In this test, we mask out portions of the ground-truth depth maps according to predefined hole-to-image (H2I) area ratios, following exisiting image inpainting approaches~\cite{wan2021high,li2022mat,ko2023continuously}. Specifically, we randomly remove depth values within predefined H2I ranges, ensuring a controlled evaluation of inpainting performance. Next, we apply each monocular depth estimators to reconstruct the occluded areas, then we recover the metric scale using Eq.~\ref{eq:least_squares}. The evaluation focused on the masked regions. From Tab.~\ref{tb:depth_inpainting} and Fig.~\ref{fig:qualitative_depth_inpainting}, we observe:

\begin{itemize}
\itemsep0mm
\item Monocular methods show stable performance across different H2I ratios, as they primarily rely on contextual cues from RGB inputs. However, they struggle to reconstruct depth accurately in small missing regions, as they do not explicitly incorporate available depth measurements.
\itemsep0mm
\item In contrast, `Proposed' consistently outperforms monocular methods across most H2I ratios, particularly when sufficient depth priors are available. For example, at H2I $\in (0.01, 0.1]$, `Proposed' achieves over 50\% lower RMSE compared with existing approaches, demonstrating its ability to effectively integrate partial depth priors and maintain global consistency.
\itemsep0mm
\item The qualitative results in Fig.~\ref{fig:qualitative_depth_inpainting} further highlight this advantage. While monocular estimators often fail to reconstruct fine structures and introduce artifacts in large missing regions, `Proposed' preserves geometric details and depth consistency.
\itemsep0mm
\item These properties also enable scale-independent learning in the refinement stage of our framework. As discussed in Sec.~\ref{subsec:deterministic_refinement}, our deterministic refinement network operates on scale-normalized depth inputs, ensuring that the model refines depth maps without being restricted to a fixed metric range. This allows our framework to generalize across datasets with varied depth distributions while still maintaining metric accuracy.
\itemsep0mm
\end{itemize}

\subsection{Analysis}
\label{subsec:analysis}
\noindent \textbf{Number of samples:}
Our uncertainty estimation relies on $\hat{\sigma}^2$ obtained from multiple diffusion samples. Fig.~\ref{fig:analysis} (a) is an example of the uncertainty map and its corresponding masked depth map as $N$ changes. We see that beyond 10 samples, additional samples provide minimal gains but increases inference time. Hence, we set $N = 10$.

\noindent \textbf{Threshold $\epsilon$:}  
Fig.~\ref{fig:analysis} (b) illustrates the masked region of a depth map given different threshold values $\epsilon$. Notably, even with a fixed $\epsilon=0.01$ most noise artifacts are effectively removed while retaining valid depth structures. This indicates that a simple thresholding approach with a fixed $\epsilon$ is sufficient for robust depth refinement.

\noindent \textbf{$\hat{\sigma}^2$ for refinement:}
To evaluate the benefit of incorporating $\hat{\sigma}^2$ into the refinement stage, we compare the refinement results with and without using $\hat{\sigma}^2$ in Fig.~\ref{fig:analysis} (c). Incorporating $\hat{\sigma}^2$ leads to more stable and accurate refinement results.

%% file: src/conclusion.tex
\section{Conclusions and Limitations}
We introduced Perfecting Depth, a novel two-stage framework for sensor depth enhancement. Unlike previous methods, our approach eliminates the need for hand-crafted priors on unreliable pixels. Instead, we employ stochastic estimation to detect artifacts and infer structural priors via a carefully designed training-inference gap, followed by deterministic refinement to ensure consistency and precision.

Despite being trained exclusively on synthetic data, our framework generalizes well to real sensor depth, demonstrating robust performance across diverse scenarios. Moreover, its training strategy prevents memorization of the metric range from training data, enabling scalability with larger and more diverse datasets. By integrating stochastic uncertainty modeling with deterministic refinement, Perfecting Depth establishes a new baseline for sensor depth enhancement.



However, our normalization process requires an additional mask for infinite-depth regions (e.g., sky); otherwise, extreme values risk compressing smaller valid depths into a narrow range, degrading performance. Incorporating insights from adaptive ordinal regression~\cite{bhat2021adabins} could further refine accuracy and enhance texture details.

%% file: src_supp/theory.tex
\section{Posterior approximation}
In this section, we will discuss about why the variation could serve as a reliable indicator for $\bD_{\text{raw}}$. \\
As shown in the main paper, the posterior at the pixel-level could be written as:
$$p(\bD_{\text{true},(i,j)} \mid \bI,\bD_{\text{cond}},\bM),$$
which can be further expanded using Bayes' rule:
\begin{equation}
\begin{aligned}
p\bigl(\bD_{\text{true},(i,j)} \mid \bI,\bD_{\text{cond}},\bM\bigr)
&\;=\;\frac{
p\bigl(\bD_{\text{cond},(i,j)} \mid \bD_{\text{true},(i,j)}\bigr)
\;p\bigl(\bD_{\text{true},(i,j)} \mid \bI,\bM\bigr)
}{
p\bigl(\bD_{\text{cond},(i,j)} \mid \bI,\bM\bigr)
}.
\end{aligned}
\end{equation}
Here we make two approximations based on the assumption that (1) the depth $\bD_{\text{cond}}$ relies more on the ground-truth depth $\bD_{\text{true}}$ at the same pixel (remember that in the training stage the synthetic data is built from the $\bD_{\text{true}}$) and (2) the RGB input $\bI$ and the missing mask $\bM$ are assumed to be independent of the ground-truth depth $\bD_{\text{true}}$ where they are fixed inputs and do not change based on the depth.  To be more specific, the RGB image \(\bI\) is a fixed input that reflects the appearance of the scene, and its generation does not explicitly depend on the depth values during inference.
The mask \(\bM\) simply indicates missing or invalid regions in the observed depth \(\bD\), and is determined by the sensor output (not the actual ground-truth depth). Thus we get:
\begin{equation}
p(\bD_{\text{cond}},\bI,\bM\mid \bD_{\text{true},(i,j)}) \approx 
p(\bD_{\text{cond}}\mid\bD_{\text{true},(i,j)}) \cdot p(\bI,\bM)
\end{equation}
Substituting the factorized likelihood back into the Bayes' Theorem, we have:

\begin{equation}
\begin{split}
p(\bD_{\text{true},(i,j)}& \mid \bI,\bD_{\text{cond}},\bM)  \approx 
\frac{p(\bD_{\text{cond},(i,j)} \mid \bD_{\text{true},(i,j)})p(\bI, \bM)p(\bD_{\text{true},(i.j)})}{p(\bD_{\text{cond}},\bI,\bM)}\\
\Rightarrow &p(\bD_{\text{true},(i,j)} \mid \bI,\bD_{\text{cond}},\bM) \propto 
p(\bD_{\text{cond},(i,j)} \mid \bD_{\text{true},(i,j)})p(\bD_{\text{true},(i.j)}).
\end{split}
\end{equation}
At this stage, we recognize that \( p(\bD_{\text{true},(i,j)}) \), the prior on the ground-truth depth, can itself depend on \( \bI \) and \( \bM \). In other words, while \( p(\bD_{\text{true},(i,j)}) \) is the prior distribution for \( \bD_{\text{true},(i,j)} \), the RGB image \( \bI \) and the mask \( \bM \) provide additional context that informs this prior. Therefore, we refine the prior term as:
$$p(\bD_{\text{true},(i,j)})=p(\bD_{\text{true},(i,j)}\mid \bI, \bM).$$
So we have:
\begin{equation}
p(\bD_{\text{true},(i,j)} \mid \bI,\bD_{\text{cond}},\bM) \propto p(\bD_{\text{cond}_(i,j)} \mid \bD_{\text{true},(i,j)})p(\bD_{\text{true},(i.j)}\mid\bI,\bM),
\end{equation}
which is exactly Eq.~8 in the main paper. As discussed in the main paper, the likelihood term $p(\bD_{\text{cond}_(i,j)} \mid \bD_{\text{true},(i,j)})$ measures how well the input depth aligns with the ground truth depth, while the prior term $p(\bD_{\text{true},(i.j)}\mid\bI,\bM)$ captures the model's learned knowledge of plausible depth values based on the RGB image and the missing mask. In real-world scenarios, when the captured depth $\bD_{\text{raw}}$ is reliable and closely matches the training data, the likelihood term becomes sharply peaked, resulting in a narrow posterior and low posterior variance. Conversely, if the captured depth is unreliable, the likelihood is weak, giving the prior term greater influence over the posterior. However, if the prior lacks sufficient information to resolve the uncertainty, the posterior becomes broader, leading to higher variance.

%% file: src_supp/training_details.tex
\section{Training Details}

\noindent\textbf{Training Dataset:}
We sample 54K RGB-D images from the Hypersim dataset~\cite{roberts2021hypersim} to train both the diffusion and refinement networks.

\noindent\textbf{Evaluation Dataset:}
For DIODE-Indoor~\cite{vasiljevic2019diode}, we use the official 325-image validation split. For NYUv2~\cite{silberman2012indoor}, we adopt the official 654-image test set. For ScanNet~\cite{dai2017scannet}, we evaluate on the 800 validation images as in Marigold~\cite{ke2024repurposing}.

\noindent\textbf{Diffusion Network:}
We fine-tune the Stable Diffusion U-Net on 4 A100 GPUs (batch size = 48) for 30K iterations, using MSE loss and a learning rate of \(\mathrm{3\times10^{-5}}\). During training, random cropping and horizontal flipping are applied to handle variability in field of view and orientation.

\noindent\textbf{Refinement Network:}
We adopt MaxViT-L~\cite{tu2022maxvit} as the encoder for the guidance module. Since we have four inputs (\(\bI\), \(\bD_{\text{rel}}\), \(\bD_{\hat{\mu}}\), and \(\hat{\sigma}^2\)), each input passes through a shallow convolutional block, and the resulting feature maps are concatenated. We also modify the first convolutional layer to match the new channel dimension. For the decoder, we follow the same architecture used in the original SDR framework~\cite{jun2024masked}.

%% file: src_supp/sensor_depth_enhancement.tex
\section{Sensor depth enhancement}
\begin{figure*}[!h]
  \centering
   \includegraphics[width=\linewidth]{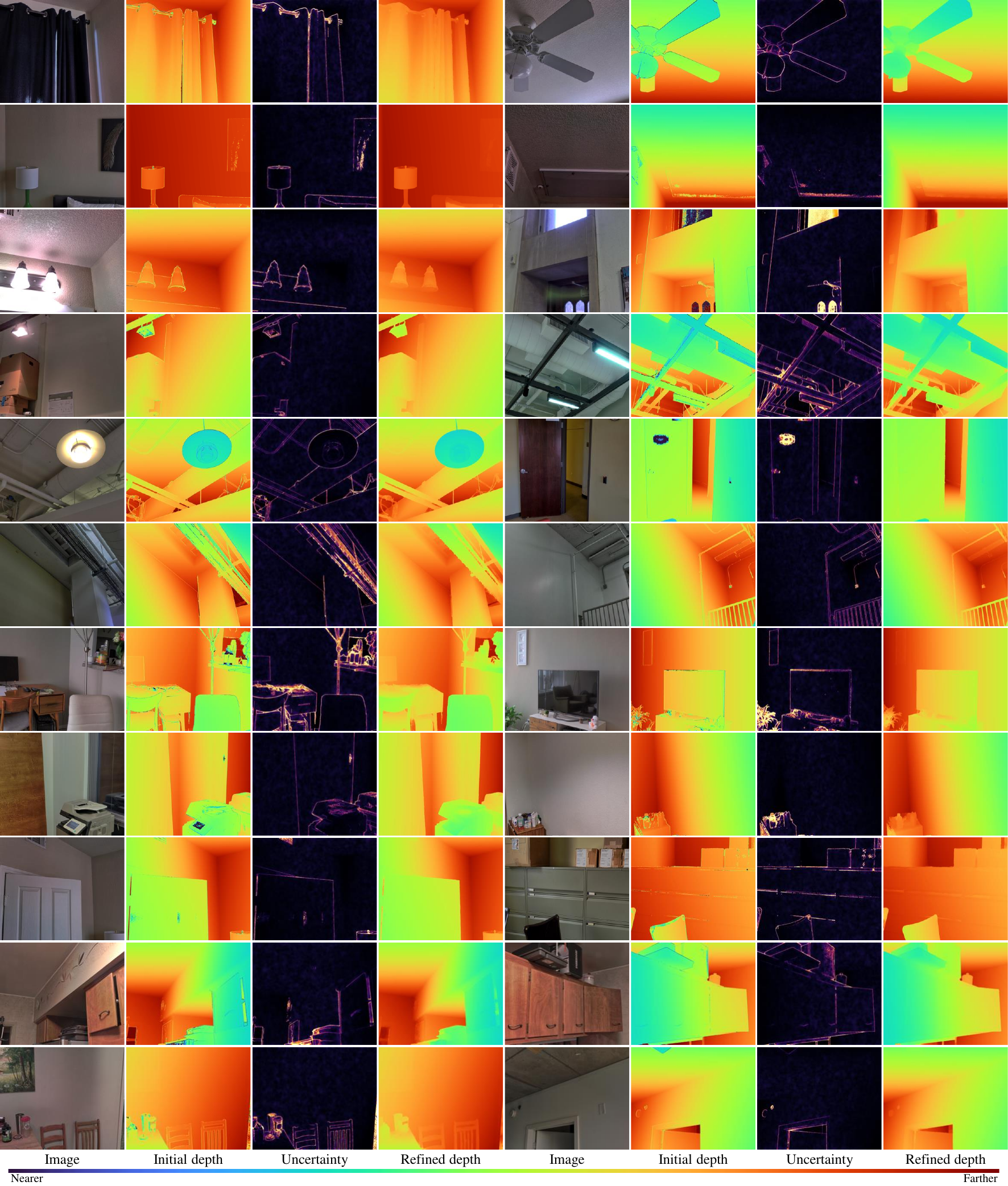}
   \caption{Sensor depth enhancement result on DIODE-Indoor.}
   \label{fig:SUPP_qualitative_sensor_depth_enh}
   \vspace{-3mm}
\end{figure*}

Fig.~\ref{fig:SUPP_qualitative_sensor_depth_enh} illustrates the sensor depth enhancement result of the proposed framework on DIODE-Indoor~\cite{vasiljevic2019diode}.

%% file: src_supp/noisy_depth_completion.tex
\section{Noisy depth completion}
\begin{figure*}[!h]
  \centering
   \includegraphics[width=\linewidth]{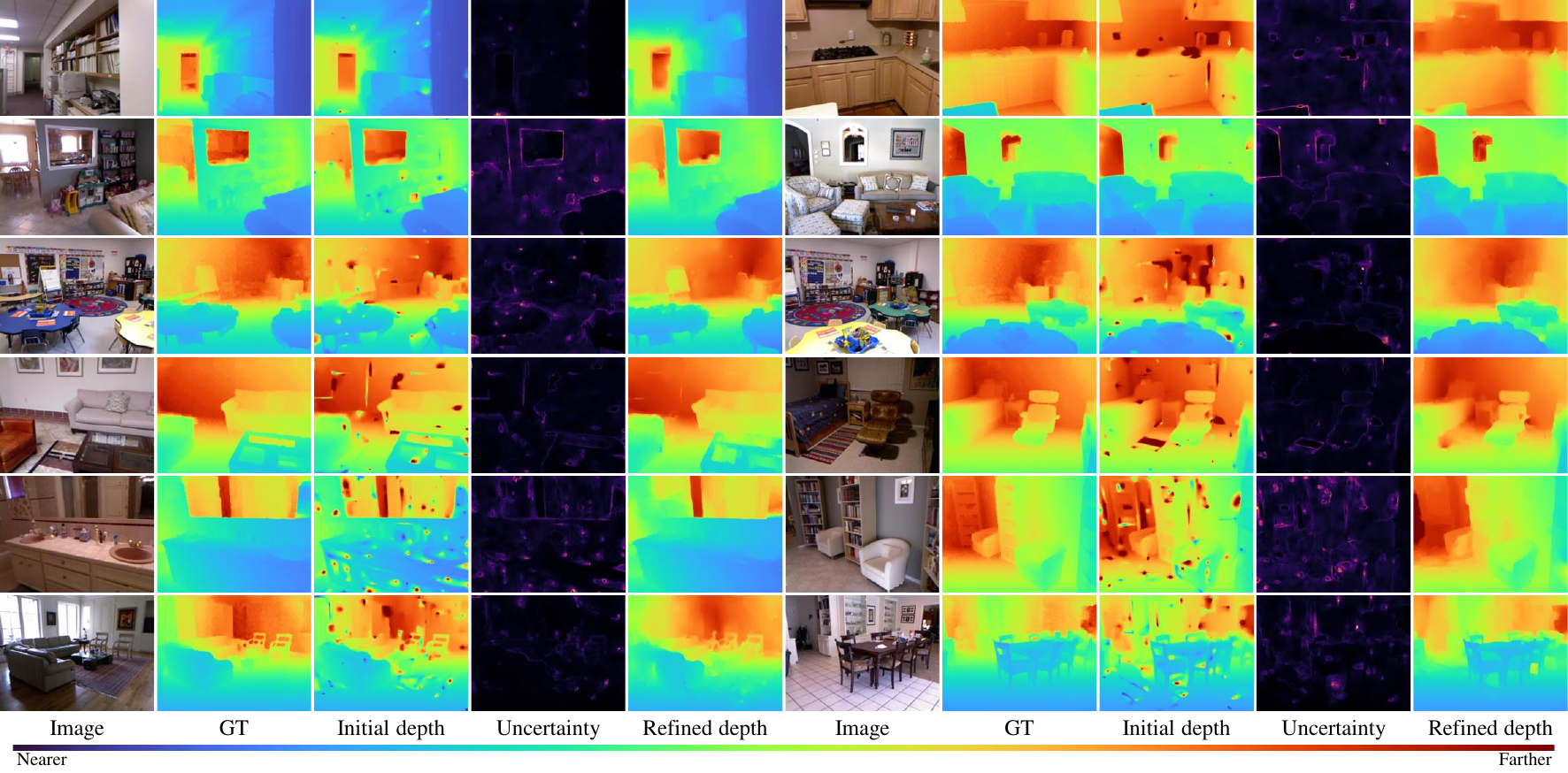}
   \caption{Refinement result on noisy depth completion using CFormer.}
   \label{fig:SUPP_qualitative_noisy_depth_completion_cformer}
   \vspace{-3mm}
\end{figure*}

\begin{figure*}[!h]
  \centering
   \includegraphics[width=\linewidth]{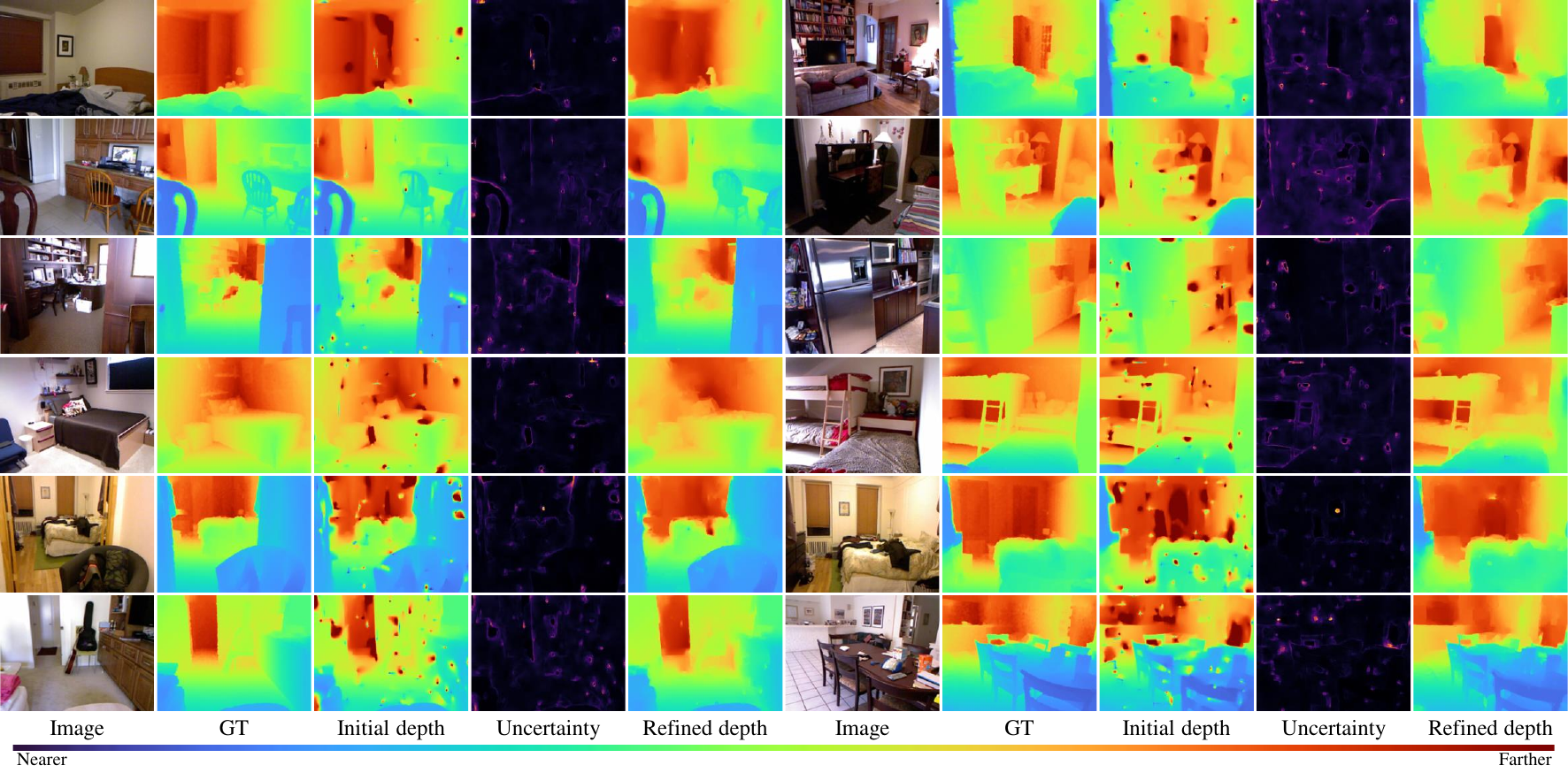}
   \caption{Refinement result on noisy depth completion using MSPN.}
   \label{fig:SUPP_qualitative_noisy_depth_completion_MSPN}
   \vspace{-3mm}
\end{figure*}

Fig.~\ref{fig:SUPP_qualitative_noisy_depth_completion_cformer} and Fig.~\ref{fig:SUPP_qualitative_noisy_depth_completion_MSPN} illustrates the refinement result of noisy depth completion on NYUv2~\cite{silberman2012indoor} using CFormer~\cite{zhang2023completionformer} and MSPN~\cite{jun2024masked}, respectively.

%% file: src_supp/depth_inpainting.tex
\section{Depth inpainting}
\begin{figure*}[!h]
  \centering
   \includegraphics[width=\linewidth]{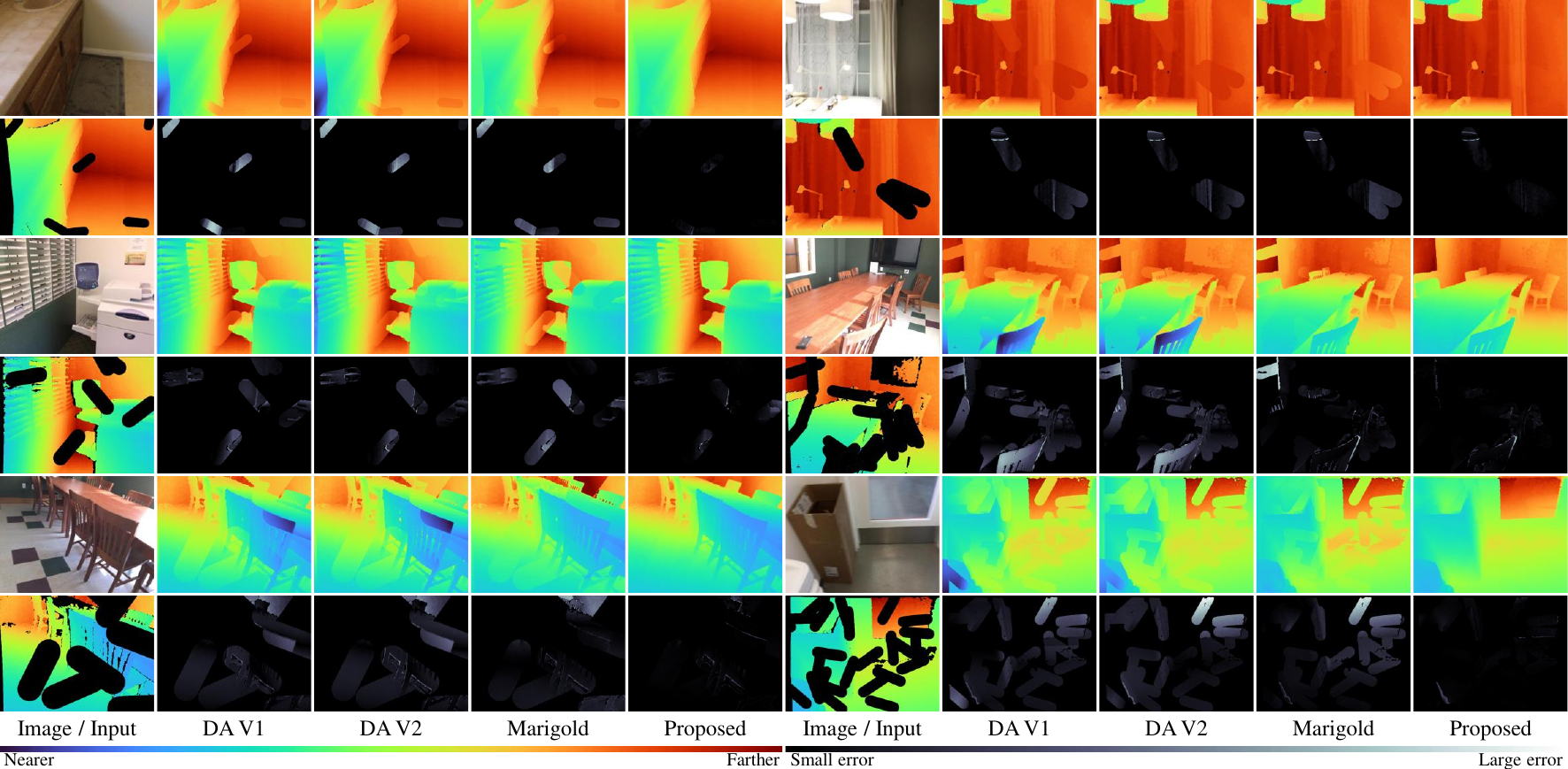}
   \caption{Qualitative comparison of depth inpainting results on ScanNet. For each depth map, the correponding error map is provided below, in which brighter pixels represent large errors.}
   \label{fig:SUPP_qualitative_depth_inpainting_scannet}
   \vspace{-3mm}
\end{figure*}

\begin{figure*}[!h]
  \centering
   \includegraphics[width=\linewidth]{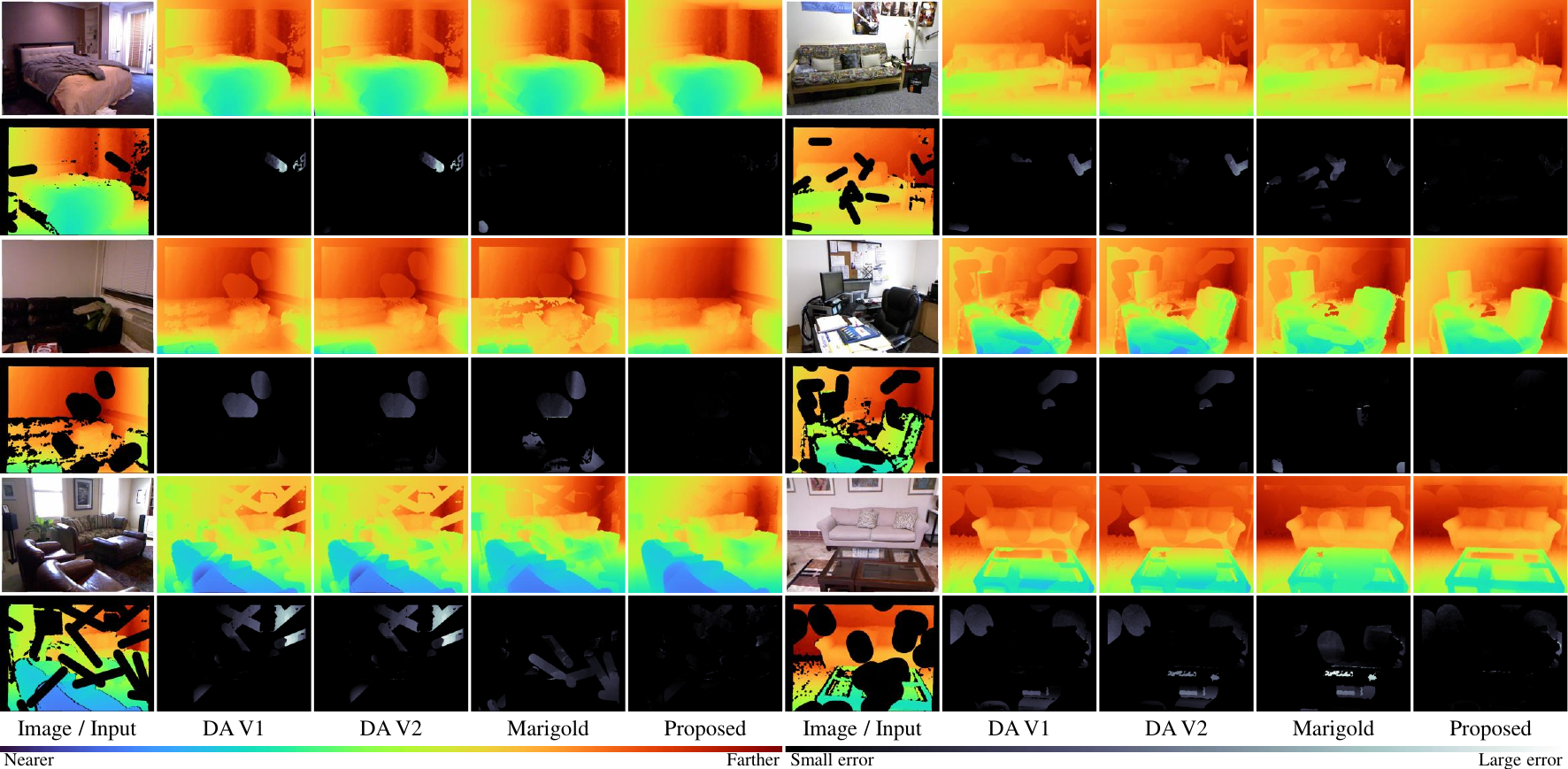}
   \caption{Qualitative comparison of depth inpainting results on NYUv2. For each depth map, the correponding error map is provided below, in which brighter pixels represent large errors.}
   \label{fig:SUPP_qualitative_depth_inpainting_nyu}
   \vspace{-3mm}
\end{figure*}

Fig.~\ref{fig:SUPP_qualitative_depth_inpainting_scannet} and Fig.~\ref{fig:SUPP_qualitative_depth_inpainting_nyu} illustrates the depth inpainting result on ScanNet~\cite{dai2017scannet} and NYUv2~\cite{silberman2012indoor}, respectively.

%% file: src_supp/ethics.tex
\section{Ethics statement}
This work is conducted as part of a research project. While we plan to share the code and findings to promote transparency and reproducibility in research, we currently have no plans to incorporate this work into a commercial product. In all aspects of this research, we are committed to adhering to Microsoft AI principles, including fairness, transparency, and accountability.